\documentclass{article}
\usepackage{spconf,amsmath,graphicx,hyperref}

\usepackage[numbers]{natbib}
\usepackage{algorithm}
\usepackage{algorithmic}

\usepackage{graphicx}
\usepackage{amsmath}
\usepackage{amssymb}
\usepackage{booktabs}
\usepackage{amsthm}

\usepackage{times}
\usepackage{epsfig}
\usepackage{fancybox,framed}
\usepackage{tabularx}
\usepackage{array}

\usepackage{verbatim}
\usepackage{multirow}

\usepackage{pifont}
\usepackage{booktabs}
\usepackage{algorithm}
\usepackage{algorithmic}

\usepackage[capitalize]{cleveref}
\crefname{section}{Sec.}{Secs.}
\Crefname{section}{Section}{Sections}
\Crefname{table}{Table}{Tables}
\crefname{table}{Tab.}{Tabs.}

\usepackage{url}

\newtheorem{mythm}{Theorem}
\newtheorem{mydef}[mythm]{Definition}

\usepackage{thm-restate} 

\newcommand{\x}{\mathbf{x}}
\newcommand{\X}{\mathbf{X}}
\newcommand{\Y}{Y}

\newcommand{\XX}{\mathcal{X}}

\newcommand{\R}{\mathbb{R}}

\title{Traj\-RS: Towards Certified Robustness in Pedestrian Trajectory Prediction}
\name{
\begin{tabular}{c}
Liang Zhang$^{1,2}$ \qquad Gaojie Jin$^{1}$ \qquad Yao Shi$^{2,4}$ \qquad Quanzhi Li$^{1}$\\
Cheng\mbox{-}Chao Huang$^{3}$ \qquad David N. Jansen$^{1}$ \qquad Lijun Zhang$^{1,2}\textsuperscript{*}$
\end{tabular}
\thanks{\textsuperscript{*}Corresponding author. This work is supported by CAS Project for Young Scientists in Basic Research, Grant No.YSBR-040, and ISCAS New Cultivation Project ISCAS-PYFX-202201.}
}
  
\address{$^{1}$Key Laboratory of System Software (Chinese Academy of Sciences) and State Key\\ 
Laboratory of Computer Science, Institute of Software, Chinese Academy of Sciences, China\\ 
$^{2}$University of Chinese Academy of Sciences,  China\\ 
$^{3}$Nanjing Institute of Software Technology, Chinese Academy of Sciences, China\\
$^{4}$Hangzhou Institute for Advanced Study, UCAS, China}
\begin{document}
\ninept
\maketitle
\begin{abstract}
The robustness of trajectory prediction models is crucial for developing safe autonomous driving systems. 
Adversarial attacks on trajectory prediction can significantly impair the accuracy of predicted trajectories, leading to hazardous driving behaviors. 
While heuristic 
defense strategies have been implemented to enhance the robustness of trajectory prediction models, these measures often fail against more sophisticated, targeted adversarial attacks. 
Hence, there is a pressing need to establish verifiable safety assurances for trajectory prediction models. 
In this paper, we extend the traditional Randomized Smoothing framework to ``Traj\-RS'', which provides a certified robust radius for smoothed trajectory predictors. 
We clarify and expand the formal definitions of robustness in trajectory prediction and tailor the practical Traj\-RS scheme specifically to ``robustness for the optimal prediction'' and ``robustness for all possible predictions''. 
An extensive set of experiments demonstrates that Traj\-RS effectively achieves robustness certification for all smoothed pedestrian trajectory predictors in this work.
\end{abstract}
\section{Introduction}
\label{sec:intro}
\begin{figure}[tb]
    \centering
    \includegraphics[width=0.95\linewidth]{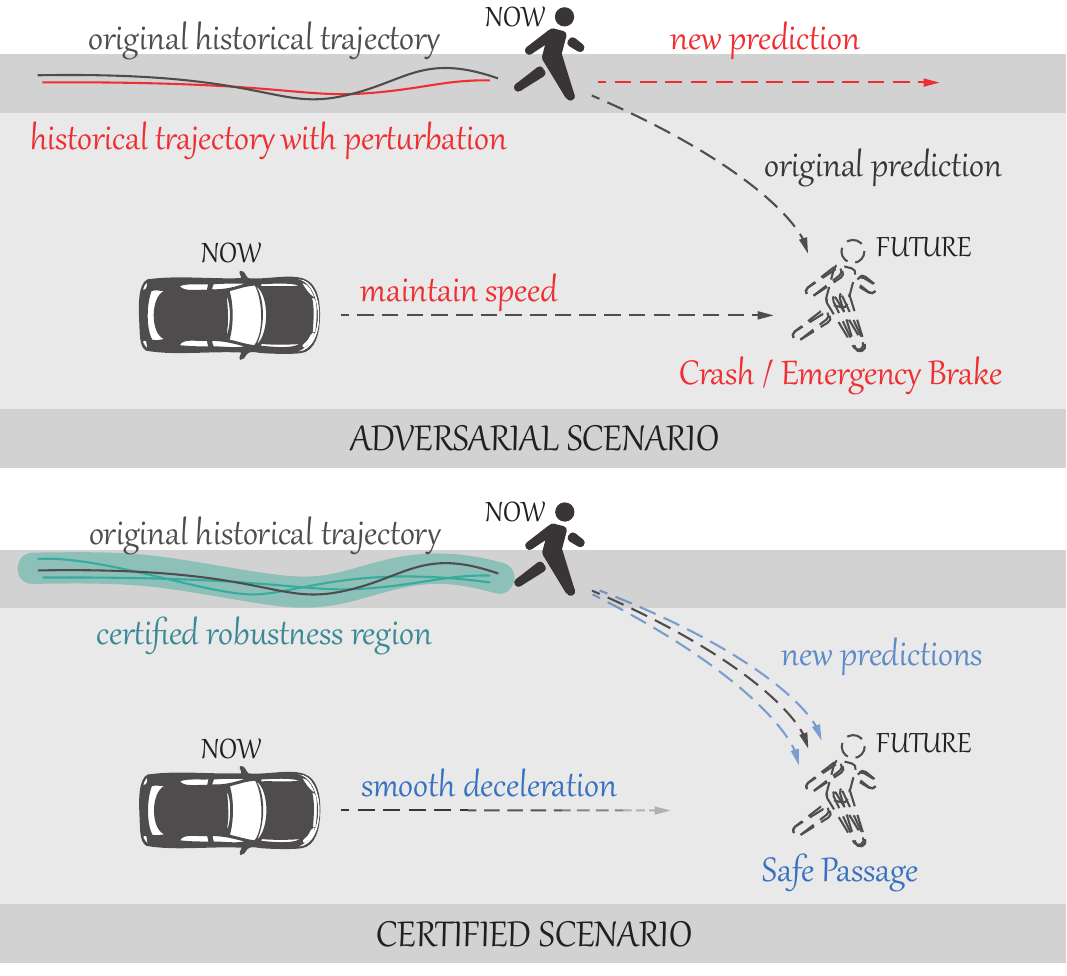}
    \vspace{-3mm}
    \caption{Two scenarios in trajectory prediction. A pedestrian prepares to cross the street. 
    In the adversarial scenario \textbf{(top),} the vehicle is misled by a disturbed historical trajectory of the pedestrian, so it mispredicts that the pedestrian will not cross the road.
    As a result, it maintains its speed, potentially causing an accident or necessitating emergency braking, endangering both the pedestrian and the vehicle.  
    In the certified scenario \textbf{(bottom),} however, any disturbances within a specified safety assurance range keep the vehicle’s prediction close to the correct future trajectory, ensuring safe traffic.}
    \label{fig:scen}
\vspace{-6mm}
\end{figure}

Trajectory prediction stands as a fundamental element in autonomous driving systems.
It is responsible for forecasting the future motion paths of nearby objects~\cite{
bae2022non,
mohamed2020social
}. 
This is essential for the vehicle to plan its forthcoming driving maneuvers safely.
While prior research has achieved success in enhancing the accuracy of trajectory prediction~\cite{DBLP:conf/cvpr/ChenZHFSWW25,DBLP:conf/cvpr/FuY0LL25,DBLP:conf/cvpr/JeongLPLY25,DBLP:conf/iclr/HuC25,DBLP:conf/iclr/0003LCL25}
, recent studies have uncovered a significant susceptibility to adversarial attacks~\cite{DBLP:conf/cvpr/YuHWLWZ25,zheng2023robustness,cao2022advdo}.
These attacks change historical trajectory data by minor perturbations,
causing major
mispredictions.
This can pose grave safety risks, as shown
in the adversarial scenario of \Cref{fig:scen} (top): the autonomous vehicle wrongly predicts that a pedestrian will stay on the sidewalk.

A few defense methods have been proposed to enhance the robustness of trajectory prediction models~\cite{zhang2022adversarial,jiao2023semi}.
Despite these accomplishments, even models previously regarded as robust have ultimately succumbed to more potent adversarial attacks, as demonstrated among others by \cite{athalye2018obfuscated}. 
This realization underscores the imperative need for methods that provide verifiable guarantees, ensuring the impregnability of the predictor against any attack within a specified perturbation radius.
This need is further amplified by the safety-critical nature of trajectory predictors in autonomous driving. 
It is only through rigorous verification that we can truly establish a safety guarantee for the model, as illustrated in the certified scenario of \Cref{fig:scen} (bottom).

To surmount the challenge, 
recent research~\cite{lecuyer2019certified, cohen2019certified, li2019certified} has introduced ``randomized smoothing''. 
This approach operates by adding smoothing noise to the input data and subsequently determining the most probable label through the smoothed classifier.
The key advantage is the ability to certify the robustness radius for the smoothed classifier. 
Randomized smoothing stands apart from other methods due to its efficiency and model-agnostic nature, making it adaptable to many varieties of models~\cite{shao2023robustness, jin2025reconcile, jin2025s, wang2025clucert}.

In this study, inspired by randomized smoothing, we extend the current trajectory prediction model to a smoothed predictor with a certified robust radius, named ``Traj\-RS''. 
Achieving this enhancement involves navigating two primary challenges. 
The first challenge pertains to the unique definition of robustness in trajectory prediction, which differs from that in image classification. 
The second challenge arises from the stochastic outputs in trajectory, necessitating robustness guarantees for the entire output distribution.

To effectively tackle these challenges, this study introduces two formal definitions of robustness for trajectory prediction.
The first, ``robustness for all possible predictions'', considers the entire output distribution.
The second, ``robustness for the optimal prediction'', concentrates on the optimal (and deterministic) predicted trajectory.
We have developed suitable randomized smoothing algorithms that implement these definitions using Monte Carlo sampling.
This approach is adept at offering robustness guarantees for the full output distribution with any given confidence and error rate.

Compared with prior verification approaches, Traj\-RS closes two key gaps. First, unlike Traj\-PAC \cite{zhang2023trajpac}—which verifies the unchanged base model via a surrogate and can at best provide a lower bound on its robust radius—Traj\-RS smooths the predictor and yields a theoretically exact certified radius for the resulting smoothed model, while also improving robustness in practice. Second, relative to the smoothing framework of Bahari et al. \cite{bahari2025certified}, which certifies only the single best output mode, Traj\-RS further certifies the entire multimodal output distribution via the smoothed model $g_A$ with controllable confidence levels and error rates.

Overall, the primary contributions of our work are as follows:

\begin{itemize}

\item We extend the Randomized Smoothing framework to Traj\-RS in \Cref{section: smoothed method} and provide a certified robust radius for smoothed trajectory predictors. 

\item We extend the formal definitions of robustness to trajectory prediction in \Cref{section: robustness} and customize the practical Traj\-RS scheme specifically to ``robustness for the optimal prediction'' and ``robustness for all possible predictions'' in \Cref{section: algorithms}.

\item We evaluate the Traj\-RS scheme against four representative trajectory forecasting models on the ETH/UCY dataset and the Stanford Drone Dataset in \Cref{section: experiments}. The experimental results demonstrate that Traj\-RS achieves effective robustness certification for all smoothed trajectory predictors in both types of robustness.
\end{itemize}

\section{Problem Formulation}

\subsection{Trajectory Prediction}

Let $\x^{(t)}\in \R^2$ be the spatial coordinate of an agent at timestamp $t$, where $\x^{(t)}=(x^{(t)}_h,x^{(t)}_v)$, $x^{(t)}_h$ represents the longitude and $x^{(t)}_v$ means the latitude.
Suppose there are $T=T_p+T_f$ timestamps, where the prior $T_p$ timestamps are situated in the past and the subsequent $T_f$ timestamps are positioned in the future. The matrix $\X_i=( \x^{(-T_p+1)}_i, \x^{(-T_p+2)}_i, \ldots, \x^{(0)}_i)\in\R^{2\times T_p}$
represents the past trajectory of the $i$-th agent, vectorized as $X_i$.
Consider $\X_0\in\R^{2\times T_p}$ is the trajectory from the to-be-predicted agent and $\mathbf{Y}_0 = ( \x^{(1)}_0, \ldots, \x^{(T_f)}_0)$ is the ground truth of the future trajectory to be predicted, vectorized as $Y_0$.
Let $\X_1,\ldots,\X_N$ be the past trajectories of the $N$ neighbouring agents.
The goal of trajectory prediction is to train a prediction model 
$f:\R^{2 T_p  (N+1)}\to \R^{2 T_f}$,
so that $f = \underset{g}{\arg\min}~\mathit{loss}(g(\XX), Y_0)$, where $\XX=\mathrm{Vec}(\X_0,\ldots,\X_N)$. $\mathrm{Vec}(\cdot)$ indicates the vectorization of the matrices.

Recent trajectory prediction models~\cite{salzmann2020trajectron++,gu2022stochastic} utilize stochastic methods to capture future movements' inherent multimodality, producing probabilistic outputs. We assume a model $f(\XX)$ outputs a discrete probability distribution over trajectories in $\R^{2T_f}$, where $\hat\Y \in f(\XX)$ represents one possible prediction.

\subsection{Robustness of Prediction Models}
\label{section: robustness}





\begin{figure}
    \centering
    \includegraphics[width=0.76\linewidth]{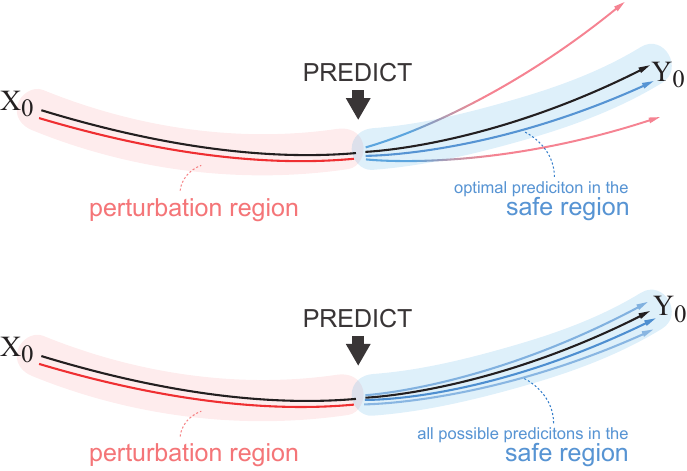}
    \vspace{-2mm}
    \caption{Two types of robustness for predictions. Robustness for the optimal prediction (\textbf{top}) concentrates on the robustness of the best-of-$k$ predicted trajectory, which is also applicable to the most probable trajectory. Robustness for all possible predictions (\textbf{bottom}) depicts the robustness of the entire output distribution. Safe region indicates the area where the distance between the predicted trajectory and the ground truth does not exceed the safety threshold $s$. Note that the perturbations are also applied to the past trajectories of the neighbouring agents, but are omitted from the figure.}
    \label{fig:robust}
    \vspace{-6mm}
\end{figure}

Building on ``label robustness'' from \cite{zhang2023trajpac}, we define ``robustness for all possible predictions'' as the ability of a model to generate safe predictions under perturbations across its entire output distribution, as illustrated in \cref{fig:robust} (bottom).

Given an input trajectory $\hat \X\in\R^{2\times T_p}$, we posit that any spatial coordinate $\x^{(t)}$ of the trajectory may be perturbed within a closed $L_2$-norm ball centered at $\x^{(t)}$ with radius $r>0$. Consequently, we define $B(\hat \X, r)$ as the set of all perturbed trajectories derived from $\hat \X$, mathematically represented as: $B(\hat \X, r)=\{\X\in\R^{2\times T_p} \mid \lVert\mathrm{Vec}(\X)-\mathrm{Vec}(\hat{\X})\rVert_2 \le r\}$.


\begin{mydef}[Robustness for All Possible Predictions] \label{def:robustness2}
Given a set of past trajectories $\hat {\XX}=\mathrm{Vec}(\hat \X_0,\hat \X_1,\ldots,\linebreak[0] \hat \X_N)$ for a target agent and its $N$ neighboring agents, and $\Y_0$ as the ground truth future trajectory of the target agent.
For a given prediction model $f$, an evaluation metric $D$, and a safety threshold $s$, model $f$ is defined to be \emph{robust for all predictions} at $\hat {\XX}$ with respect to a perturbation radius $r > 0$ if:
for any $\XX = \mathrm{Vec}(\X_0,\X_1,\ldots,\linebreak[0]\X_N)$, where $\X_i\in B(\hat \X_i,r)$ and any $\Y \in f(\XX)$,
it satisfies $D(\Y, \Y_0) \leq s$.
\end{mydef}

We utilize Average Displacement Error (ADE) and Final Displacement Error (FDE)~\cite{alahi2014socially,alahi2016social,gupta2018social,li2019conditional} as the evaluation metric $D$. $L_2$-norm for the assessment of robustness provides an intuitive geometric interpretation of perturbations in trajectory space.

However, this definition has its limitations. Many recent methods~\cite{yuan2021agentformer} explicitly encourage the prediction of a diverse set of trajectories that span different directions. For these models, the required safety distance $s$ to validate robustness would be significantly larger. Moreover, most adversarial robustness methods~\cite{cao2022advdo} focus on best-of-$k$ predictions under attack, aligning with accuracy metrics commonly used in recent studies~\cite{alahi2014socially,alahi2016social,gupta2018social,li2019conditional}.

In light of these considerations, we propose a second definition of robustness, as illustrated in \cref{fig:robust} (top):

\begin{mydef}[Robustness for the Optimal Prediction]  \label{def:robustness1}
Given a set of past trajectories $\hat {\XX}=\mathrm{Vec}(\hat \X_0,\hat \X_1,\ldots,\linebreak[0] \hat \X_N)$ for a target agent and its $N$ neighboring agents, and $\Y_0$ as the ground truth of future trajectory for the target agent. 
For a given prediction model $f$, an evaluation metric $D$, and a predefined safety threshold $s$, model $f$ is defined to be \emph{robust for the optimal prediction} at $\hat {\XX}$ with respect to a perturbation radius $r > 0$ if: 
for any $\XX = \mathrm{Vec}(\X_0,\X_1,\ldots,\linebreak[0]\X_N)$, where $\X_i\in B(\hat \X_i,r)$, 
the trajectory $\Y_{opt} = \underset{Y \in f(\XX)}{\arg\min}~D(Y, Y_0)$
satisfies $D(\Y_{opt}, \Y_0) \leq s$.
\end{mydef}


\section{Method}

\subsection{Smoothed Method and Robustness Guarantee}
\label{section: smoothed method}

Randomized smoothing~\cite{cohen2019certified} is a black-box technique that converts a base function $f$ into a certifiably robust smoothed function $g$ by injecting Gaussian noise into the input $x$ and returning the most likely output of $f$. 
For the two robustness notions in \cref{section: robustness}, we define two smoothed models \(g_O\) (optimal prediction) and \(g_A\) (all predictions). 
Let \(\XX=\mathrm{Vec}(\X_0,\ldots,\X_N)\), \(\varepsilon\sim\mathcal{N}(0,\sigma^2 I)\), distance metric \(D\), ground truth \(\Y_0\), and safety threshold \(s\).
Define the events
\[
\mathcal{E}_O:\; D(\Y_{\mathrm{opt}},\Y_0)\le s,\quad 
\Y_{\mathrm{opt}}=\arg\min_{Y\in f(\XX+\varepsilon)} D(Y,\Y_0),
\]
\[
\mathcal{E}_A:\; \forall\, Y\in f(\XX+\varepsilon),\; D(Y,\Y_0)\le s.
\]
Let \(p_m=\mathbb{P}(\mathcal{E}_m)\) for \(m\in\{O,A\}\). We set

\begin{equation}
\label{eq:1}
g_m(\XX)=\begin{cases}
1,&\text{if } p_m\ge \tfrac{1}{2},\\
0,&\text{otherwise.}
\end{cases}
\vspace{-3mm}
\end{equation}

\begin{restatable}[Robustness Guarantee]{mythm}{unifiedrobustness}
\label{thm:unified}
For \(m\in\{O,A\}\), suppose there exists \(\underline{p}_m\in(\tfrac{1}{2},1]\) such that:
\begin{equation}
\label{eq:2}
p_m=\mathbb{P}(\mathcal{E}_m)\ge \underline{p}_m.
\end{equation}
Then, for all perturbations \(\delta\) with \(\lVert\delta\rVert_2\le R_m\), we have \(g_m(\XX+\delta)=1\), where
\begin{equation}
\label{eq:3}
R_m=\sigma\,\Phi^{-1}(\underline{p}_m),
\end{equation}
and \(\Phi^{-1}\) is the standard normal quantile. 
\end{restatable}

In practice we estimate a one-sided lower confidence bound \(\underline{p}_m\) (via binomial inference), and all certificates remain valid when \(p_m\) is replaced by this lower bound. 
We may find out that: leveraging randomized smoothing’s black-box nature, our smoothing approach applies universally to prediction models, furnishing precise robustness guarantees.
Note that Gaussian noise is used during smoothing and influences how the certifiable robust radius is calculated (\cref{eq:3}), but the resulting smoothed model is robust against \textbf{any perturbation} within the certifiable robust radius, regardless of distribution.

\subsection{Practical Algorithm}
\label{section: algorithms}

We evaluate $g_O(\XX)$ and $g_A(\XX)$ with a single Monte Carlo procedure that estimates a one-sided $(1-\alpha)$ lower bound $\underline{p}$ on the safety probability and maps it to a certified radius $R=\sigma\,\Phi^{-1}(\underline{p})$, as shown in \Cref{A:1}. The two variants differ in (i) how the representative prediction is chosen and (ii) the certification statistic.

\begin{algorithm}[!ht]
    \small
	\renewcommand{\algorithmicrequire}{\textbf{Input:}}
	\renewcommand{\algorithmicensure}{\textbf{Output:}}
    \newcommand{\algorithmicprediction}{\textbf{Prediction Phase:}}
    \newcommand{\algorithmiccertification}{\textbf{Certification Phase:}}
    \newcommand{\PREDICTION}{\item[\algorithmicprediction]}
    \newcommand{\CERTIFICATION}{\item[\algorithmiccertification]}
	\caption{Evaluation \& certification for $g_O$ and $g_A$}
	\label{alg:unified}
	\begin{algorithmic}[1]
        \REQUIRE Past trajectories $\XX=(\X_0,\ldots,\X_N)$, ground-truth $\Y_0$, safety threshold $s$; base predictor $f$; number of perturbations $n$; Gaussian noise std.\ $\sigma$; significance $\alpha$; evaluation metric $D$ (e.g., ADE); \textbf{mode} $m\!\in\!\{\mathsf{O},\mathsf{A}\}$.
        \REQUIRE If $m{=}\mathsf{O}$: best-of-$k$ size $k$. \quad If $m{=}\mathsf{A}$: per-input draws $n_p$ and target per-input safety level $\tau$ (default $\tau{=}0.99$).
        \ENSURE Prediction $\widehat{\Y}$ and certified radius $R$ (or ``cannot certify'').
        \PREDICTION
        \STATE For $i{=}1{:}n$: sample $\varepsilon_i \sim \mathcal{N}(0,\sigma^2 I)$ and set $\XX_i=\XX+\varepsilon_i$.
        \STATE \textbf{If} $m{=}\mathsf{O}$ \textbf{then} generate $k$ predictions $\{\Y_{ij}\}_{j=1}^k \leftarrow f(\XX_i)$; \textbf{else} ($m{=}\mathsf{A}$) generate $n_p$ predictions $\{\Y_{ij}\}_{j=1}^{n_p} \leftarrow f(\XX_i)$.
        \STATE Let $\mathcal{S}=\{\Y_{ij}\}$ over all $i$ and $j$.
        \STATE \textbf{Representative selection:}
        \STATE \quad \textbf{If} $m{=}\mathsf{O}$: cluster $\mathcal{S}$ into $k$ groups by K-medoids (K=$k$); set $\widehat{\Y}$ to the cluster medoid whose medoid is closest to $\Y_0$ under $D$ (best-of-$k$ among medoids).
        \STATE \quad \textbf{If} $m{=}\mathsf{A}$: compute the K-medoids medoid of $\mathcal{S}$ with K=$1$ and set $\widehat{\Y}$ to this medoid.
        \CERTIFICATION
        \STATE \textbf{If} $m{=}\mathsf{O}$:
        \STATE \quad For each $i$, let $\Y^{\mathrm{opt}}_i=\arg\min_j D(\Y_{ij},\Y_0)$; set $n_{\mathrm{safe}}=\sum_{i=1}^n \mathbb{I}\!\left(D(\Y^{\mathrm{opt}}_i,\Y_0)\le s\right)$.
        \STATE \quad $\underline{p}=\mathrm{LOWERCONFBOUND}(n_{\mathrm{safe}}, n, 1{-}\alpha)$.
        \STATE \textbf{Else} ($m{=}\mathsf{A}$):
        \STATE \quad For each $i$, let $c_i=\sum_{j}\mathbb{I}\!\left(D(\Y_{ij},\Y_0)\le s\right)$ and $q_i=\mathrm{LOWERCONFBOUND}(c_i,\, n_p,\, 1{-}\alpha)$.
        \STATE \quad Mark $i$ safe if $q_i \ge \tau$; let $n_{\mathrm{safe}}=\sum_{i=1}^n \mathbb{I}(q_i \ge \tau)$; set $\underline{p}=\mathrm{LOWERCONFBOUND}(n_{\mathrm{safe}}, n, 1{-}\alpha)$.
        \STATE \textbf{If} $\underline{p}\le \tfrac{1}{2}$ \textbf{or} $D(\widehat{\Y},\Y_0)>s$ \textbf{then} return ``cannot certify''; \textbf{else} return $\widehat{\Y}$ and $R=\sigma\,\Phi^{-1}(\underline{p})$.
	\end{algorithmic}
\label{A:1}
\end{algorithm}

\noindent\textbf{Notes.} We use K\mbox{-}medoids for representative selection to remain robust to outliers and to avoid averaging artifacts that may distort physically plausible trajectories (compared to K\mbox{-}means)~\cite{velmurugan2010computational}. $\mathrm{LOWERCONFBOUND}$ denotes a one-sided $(1{-}\alpha)$ binomial lower confidence bound (e.g., Clopper--Pearson), and $\Phi^{-1}$ is the standard normal quantile.

\section{Experiments}
\label{section: experiments}

We evaluate Traj\-RS on two benchmarks: ETH/\linebreak[0]UCY~\cite{DBLP:conf/iccv/PellegriniESG09,lerner2007crowds} and Stanford Drone Dataset (SDD)~\cite{robicquet2016learning}, using $8$ past steps ($0.4$s each) to predict the next $12$~\cite{salzmann2020trajectron++}. Experiments cover four representative models: Trajectron++~\cite{salzmann2020trajectron++}, AgentFormer~\cite{yuan2021agentformer}, MemoNet~\cite{xu2022remember}, and MID~\cite{gu2022stochastic}. The most commonly used evaluation metrics—Average Displacement Error (ADE) and Final Displacement Error (FDE)~\cite{alahi2014socially,alahi2016social,gupta2018social,li2019conditional}—are employed with a best-of-$k$ minimum ADE/FDE setting ($k{=}20$)~\cite{gupta2018social,sadeghian2019sophie,salzmann2020trajectron++,phan2020covernet}. 
Due to space, we report ADE in the main text; across all settings, FDE exhibits the same trends as ADE.

\subsection{Robustness for the Optimal Prediction}

We evaluate the pre-trained models on a randomly selected subset of $500$ samples from both the ETH/UCY and SDD datasets, due to the time required for sampling. To account for unit differences (meters vs.\ pixels), we apply dataset-specific noise levels: $\sigma \in \{0.1, 0.4, 0.7, 1.0\}$ for ETH/UCY and $\sigma \in \{1, 4, 7, 10\}$ for SDD, with safety thresholds of $s=2$ (ETH/UCY) and $s=50$ (SDD). Each model uses $n=10^4$ Monte Carlo samples with significance level $\alpha = 0.001$. 

\begin{figure}[t]
\includegraphics[width=1\linewidth]{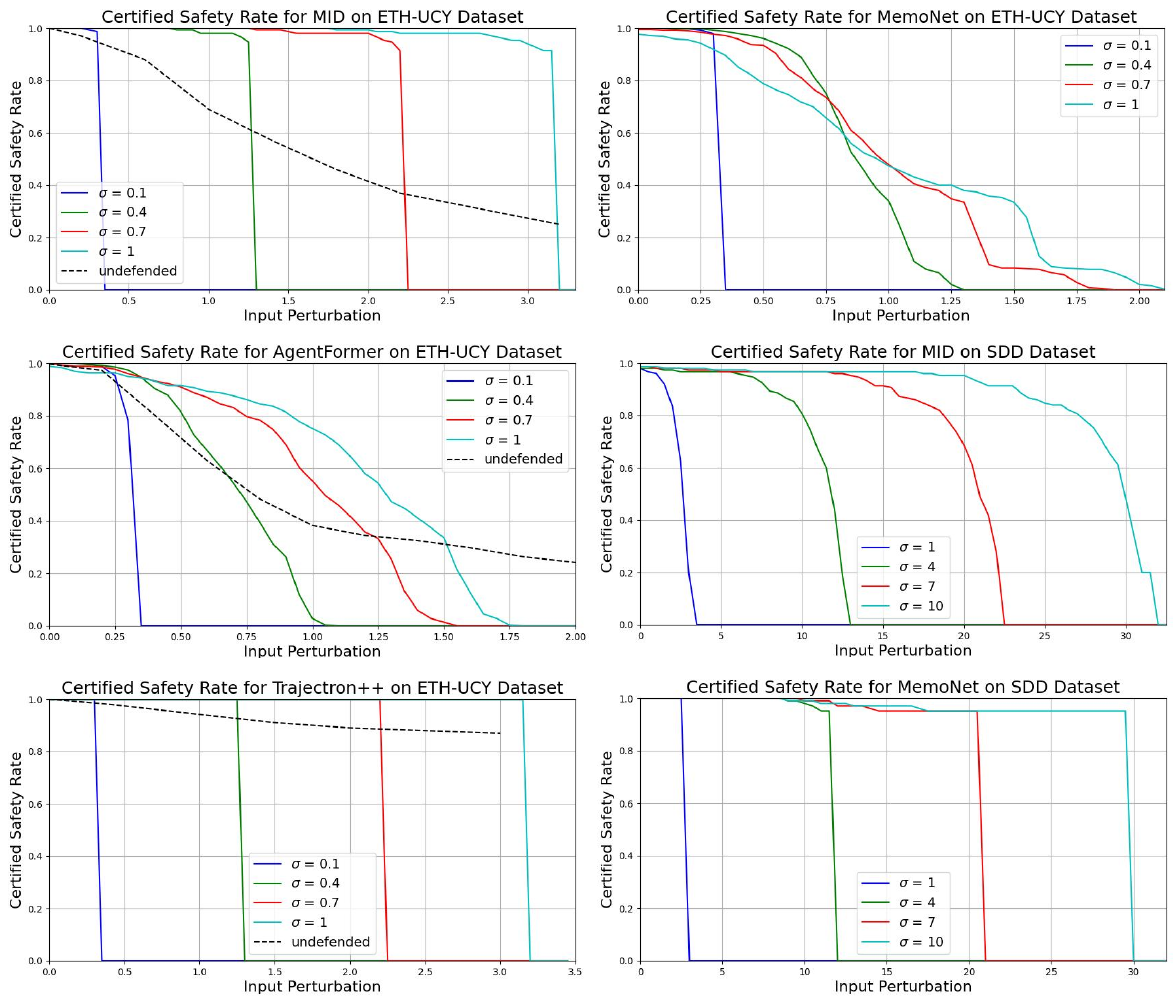}
\vspace{-8mm}
\caption{Certified safety rates (ADE metric) for smoothed MID, AgentFormer, Trajectron++,  and MemoNet on the ETH/UCY dataset, and smoothed MID and MemoNet on the SDD dataset.}
\label{fig:6-pic}
\vspace{-1mm}
\end{figure}

\begin{table}[tb]
\centering
\setlength{\tabcolsep}{1.2mm}
\scalebox{1}{
\begin{tabular}{clccccc}
\specialrule{.1em}{.0em}{.0em}
ETH/UCY    &  & Base model & $\sigma{=}0.1$   & $\sigma{=}0.4$   & $\sigma{=}0.7$   & $\sigma{=}1.0$   \\ \cline{1-1} \cline{3-7} 
Traj++     &  & 0.29       & \textbf{0.28} & \textbf{0.27} & \textbf{0.27} & \textbf{0.26} \\
AgentFormer &  & 0.28       & \textbf{0.14} & \textbf{0.26}      & 0.35          & 0.41          \\
MID         &  & 0.27       & \textbf{0.21} & \textbf{0.21} & \textbf{0.22} & \textbf{0.21} \\
MemoNet     &  & 0.21       & 0.23          & 0.39          & 0.54          & 0.59          \\
\specialrule{.1em}{.0em}{.0em}
SDD         &  & Base model & $\sigma{=}1$     & $\sigma{=}4$     & $\sigma{=}7$     & $\sigma{=}10$    \\ \cline{1-1} \cline{3-7} 
MID         &  & 5.87       & 8.56          & 7.06 & 6.17 & 5.94 \\
MemoNet     &  & 5.58       & 6.34 & 5.97 & 6.62 & 6.75 \\
\specialrule{.1em}{.0em}{.0em}
\end{tabular}
}
\vspace{-3mm}
\caption{Quantitative prediction results for different base models and their smoothed models (several noise levels) on the ETH/UCY (\textbf{top}) and SDD (\textbf{bottom}) datasets with best-of-20 strategy in ADE metric. Lower is better.}
\label{table: output}
\vspace{-5mm}
\end{table}

\textbf{Prediction performance.}
\Cref{table: output} compares ADE of smoothed models at different noise levels with base models. Moderate noise ($\sigma$) slightly affects or even improves accuracy for Trajectron++ and MID; larger noise degrades AgentFormer and MemoNet, reflecting an accuracy–robustness trade-off. The gains at low–medium noise likely stem from the clustering operation, which reduces the randomness of predictions that deviate significantly from the ground truth.

\textbf{Certification performance.}
We define the Certified Safety Rate (CSR), analogous to certified accuracy~\cite{cohen2019certified}, as the proportion of test samples safely predicted within a certifiably robust $L_2$ radius $r$:

\vspace{-5mm}
\begin{small}
\begin{equation}
\label{eq:7}
\begin{aligned}
\mathrm{CSR}(r)=\frac{1}{n} \sum_{i=1}^n[\mathbb{I}(\mathrm{CR}(\XX_i) &\geqslant r) \mathbb{I}(\mathrm{D}(\XX_i)\leqslant s)],
\\ \XX_i  \in\left(\XX_1,\ldots,\XX_n\right),
\end{aligned}
\vspace{-1mm}
\end{equation}
\end{small}
where $\mathrm{CR}(\XX_i)$ is the certified radius of model $g$ in $\XX_i$ and $\mathrm{D}$ is the minimum ADE/FDE in~\cite{gupta2018social,sadeghian2019sophie,salzmann2020trajectron++,phan2020covernet}, with safety threshold $s$.

\Cref{fig:6-pic} shows CSR (ADE) curves that decline gradually with radius $r$ and then drop sharply at the certification limit determined by $\sigma$ and $n$~\cite{cohen2019certified}. We also apply PGD attacks~\cite{madry2017towards} to estimate empirical CSR upper bounds for the original models, depicted as dashed black lines
in \Cref{fig:6-pic} (left). Smoothed models remain above these bounds until the sharp drop.

\noindent\doublebox{
\begin{minipage}{0.94\linewidth}
{\bf Conclusion 1:}
Smoothing a model does enhance its robustness in our case studies, measured as CSR.
It does not lead to a severe decline in prediction accuracy, measured as ADE/FDE.
\end{minipage}
}

Each subplot in \cref{fig:6-pic} illustrates that higher noise levels ($\sigma$) lead to larger certifiable radii, with minor accuracy drops observed at smaller radii across different smoothed models.
Additional ablation studies on the parameters $s$, $\alpha$, and $n$ show that varying $s$  does not affect the observed robustness–accuracy trade-off or the advantage of smoothing. CSR is largely insensitive to $\alpha$. Larger $n$ primarily extends the maximal certifiable radius. As for efficiency, Traj\-RS is competitive with mainstream verification methods like~\cite{zhang2023trajpac}.

\noindent\doublebox{
\begin{minipage}{0.94\linewidth}
{\bf Conclusion 2:} Traj\-RS demonstrates scalable verification of robustness in various prediction models, offering a flexible trade-off between accuracy and robustness by adjusting noise levels. The adaptability ensures it meets verification demands for varying confidence levels and safety thresholds in different scenarios.
\end{minipage}
}

\begin{table}[tb]
\centering
\setlength{\tabcolsep}{0.12mm}
\scalebox{1}{
\begin{tabular}{cccccccccc}
\specialrule{.1em}{.0em}{.0em} 
Scene    &  & ID          &  & Smoothed Model &  & $s=2$ & $s=3$ & $s=4$                     & $s=5$                      \\ \cline{1-1} \cline{3-3} \cline{5-5} \cline{7-10} 
         &  &             &  & Trajectron++   &  & 0.73 & 0.73 & 0.73                     & 0.73                     \\
hotel    &  & (7550, 157) &  & AgentFormer    &  & -    & 0.25 & 0.48                     & 0.70                     \\
         &  &             &  & MID            &  & 0.73 & 0.73 & 0.73                     & 0.73                     \\
          &  &             &  & MemoNet        &  & -    & 0.17 & 0.38                     & 0.56                   \\ \cline{1-1} \cline{3-3} \cline{5-5} \cline{7-10} 
 \specialrule{.05em}{.0em}{.0em}
 Scene    &  & ID          &  & Smoothed Model &  & $s=40$ & $s=50$ & $s=60$                     & $s=70$                     \\ \cline{1-1} \cline{3-3} \cline{5-5} \cline{7-10} 
 quad\_0  &  & (84, 5)     &  & MID            &  & 4.39 & 6.60 & 7.29 & 7.29 \\ \cline{1-1} \cline{3-3} \cline{5-5} \cline{7-10} 
\specialrule{.08em}{.0em}{.0em}
\end{tabular}
}
\vspace{-3mm}
\caption{Certified robustness radii for various smoothed models (ADE metric) under the criterion of ``robustness for all possible predictions'' on samples from ETH/UCY (\textbf{top}) and SDD (\textbf{bottom}). A ``-'' denotes the model's inability to provide a certified robustness radius in the given context.}
\label{table: ALL}
\vspace{-4mm}
\end{table}

\subsection{Robustness for All Possible Predictions}
\label{subsec: all possible}

Following the setup in \cite{zhang2023trajpac}, we selected specific trajectories from ETH/UCY (Trajectron++, MemoNet, MID, AgentFormer; $\sigma=0.3$) and SDD (MID; $\sigma=3$), identified by (frame ID, person ID). We used $n=n_p=1000$ and significance level $\alpha=0.001$.

\Cref{table: ALL} summarizes ADE results across various safety thresholds.
Traj\-RS successfully provides certifiable robustness radii, which generally increase with higher safety thresholds until reaching theoretical limits determined by noise level ($\sigma$) and sample size ($n$). Conversely, lower safety thresholds present certification challenges due to the multimodal nature of predictions and may require larger sample sizes. These results suggest that validating robustness for all possible predictions is more effectively conducted under scenarios with higher safety thresholds.

\noindent\doublebox{
\begin{minipage}{0.94\linewidth}
{\bf Conclusion 3:} When the required safety distance is not particularly small, our method is highly effective at verifying robustness for all possible predictions across different models, providing the corresponding robust radius for each model.
\end{minipage}
}
\section{Conclusion}

In this study, we extend the formal notion of robustness to trajectory prediction and propose Traj\-RS, a certifiable framework for smoothing-based robust prediction. Traj\-RS enhances robustness of existing models, offering certified guarantees under input perturbations. Extensive experiments validate its effectiveness. Future work will explore applying Traj\-RS to improve the robustness and safety of autonomous vehicles in real-world scenarios.


\vfill\pagebreak

\bibliographystyle{IEEEbib}
\bibliography{strings,refs.cleaned}

@String(CVPR  = {IEEE Conf. Comput. Vis. Pattern Recog.})

@String(ICCV  = {Int. Conf. Comput. Vis.})

@String(ICLR  = {Int. Conf. Learn. Represent.})

@String(CVPR  = {CVPR})

@String(ICCV  = {ICCV})

@String(ICLR  = {ICLR})

@inproceedings{jiao2023semi,
  title={Semi-supervised Semantics-guided Adversarial Training for Robust Trajectory Prediction},
  author={Jiao, Ruochen and Liu, Xiangguo and Sato, Takami and Chen, Qi Alfred and Zhu, Qi},
  booktitle={Proceedings of the IEEE/CVF International Conference on Computer Vision},
  pages={8207--8217},
  year={2023}
}

@inproceedings{shao2023robustness,
  title={Robustness Certification for Structured Prediction with General Inputs via Safe Region Modeling in the Semimetric Output Space},
  author={Shao, Huaqing and Wang, Lanjun and Yan, Junchi},
  booktitle={Proceedings of the 29th ACM SIGKDD Conference on Knowledge Discovery and Data Mining},
  pages={2010--2022},
  year={2023}
}

@inproceedings{zhang2023trajpac,
  title={TrajPAC: Towards Robustness Verification of Pedestrian Trajectory Prediction Models},
  author={Zhang, Liang and Xu, Nathaniel and Yang, Pengfei and Jin, Gaojie and Huang, Cheng-Chao and Zhang, Lijun},
  booktitle={Proceedings of the IEEE/CVF International Conference on Computer Vision},
  pages={8327--8339},
  year={2023}
}

@inproceedings{bae2022non,
  title={Non-probability sampling network for stochastic human trajectory prediction},
  author={Bae, Inhwan and Park, Jin-Hwi and Jeon, Hae-Gon},
  booktitle={Proceedings of the IEEE/CVF Conference on Computer Vision and Pattern Recognition},
  pages={6477--6487},
  year={2022}
}

@inproceedings{alahi2016social,
  title={Social lstm: Human trajectory prediction in crowded spaces},
  author={Alahi, Alexandre and Goel, Kratarth and Ramanathan, Vignesh and Robicquet, Alexandre and Fei-Fei, Li and Savarese, Silvio},
  booktitle={Proceedings of the IEEE conference on computer vision and pattern recognition},
  pages={961--971},
  year={2016}
}

@inproceedings{mohamed2020social,
  title={Social-stgcnn: A social spatio-temporal graph convolutional neural network for human trajectory prediction},
  author={Mohamed, Abduallah and Qian, Kun and Elhoseiny, Mohamed and Claudel, Christian},
  booktitle={Proceedings of the IEEE/CVF conference on computer vision and pattern recognition},
  pages={14424--14432},
  year={2020}
}

@inproceedings{gu2022stochastic,
  title={Stochastic trajectory prediction via motion indeterminacy diffusion},
  author={Gu, Tianpei and Chen, Guangyi and Li, Junlong and Lin, Chunze and Rao, Yongming and Zhou, Jie and Lu, Jiwen},
  booktitle={Proceedings of the IEEE/CVF Conference on Computer Vision and Pattern Recognition},
  pages={17113--17122},
  year={2022}
}

@inproceedings{gupta2018social,
  title={Social gan: Socially acceptable trajectories with generative adversarial networks},
  author={Gupta, Agrim and Johnson, Justin and Fei-Fei, Li and Savarese, Silvio and Alahi, Alexandre},
  booktitle={Proceedings of the IEEE conference on computer vision and pattern recognition},
  pages={2255--2264},
  year={2018}
}

@inproceedings{sadeghian2019sophie,
  title={Sophie: An attentive gan for predicting paths compliant to social and physical constraints},
  author={Sadeghian, Amir and Kosaraju, Vineet and Sadeghian, Ali and Hirose, Noriaki and Rezatofighi, Hamid and Savarese, Silvio},
  booktitle={Proceedings of the IEEE/CVF conference on computer vision and pattern recognition},
  pages={1349--1358},
  year={2019}
}

@inproceedings{li2019conditional,
  title={Conditional generative neural system for probabilistic trajectory prediction},
  author={Li, Jiachen and Ma, Hengbo and Tomizuka, Masayoshi},
  booktitle={2019 IEEE/RSJ International Conference on Intelligent Robots and Systems (IROS)},
  pages={6150--6156},
  year={2019},
  publisher={IEEE}
}

@inproceedings{salzmann2020trajectron++,
  title={Trajectron++: Dynamically-feasible trajectory forecasting with heterogeneous data},
  author={Salzmann, Tim and Ivanovic, Boris and Chakravarty, Punarjay and Pavone, Marco},
  booktitle={Computer Vision--ECCV 2020: 16th European Conference, Glasgow, UK, August 23--28, 2020, Proceedings, Part XVIII 16},
  pages={683--700},
  year={2020},
  publisher={Springer}
}

@inproceedings{yuan2021agentformer,
  title={Agentformer: Agent-aware transformers for socio-temporal multi-agent forecasting},
  author={Yuan, Ye and Weng, Xinshuo and Ou, Yanglan and Kitani, Kris M},
  booktitle={Proceedings of the IEEE/CVF International Conference on Computer Vision},
  pages={9813--9823},
  year={2021}
}

@inproceedings{xu2022remember,
  title={Remember intentions: retrospective-memory-based trajectory prediction},
  author={Xu, Chenxin and Mao, Weibo and Zhang, Wenjun and Chen, Siheng},
  booktitle={Proceedings of the IEEE/CVF Conference on Computer Vision and Pattern Recognition},
  pages={6488--6497},
  year={2022}
}

@inproceedings{zhang2022adversarial,
  title={On adversarial robustness of trajectory prediction for autonomous vehicles},
  author={Zhang, Qingzhao and Hu, Shengtuo and Sun, Jiachen and Chen, Qi Alfred and Mao, Z Morley},
  booktitle={Proceedings of the IEEE/CVF Conference on Computer Vision and Pattern Recognition},
  pages={15159--15168},
  year={2022}
}

@inproceedings{cao2022advdo,
  title={Advdo: Realistic adversarial attacks for trajectory prediction},
  author={Cao, Yulong and Xiao, Chaowei and Anandkumar, Anima and Xu, Danfei and Pavone, Marco},
  booktitle={Computer Vision--ECCV 2022: 17th European Conference, Tel Aviv, Israel, October 23--27, 2022, Proceedings, Part V},
  pages={36--52},
  year={2022},
  organization={Springer}
}

@inproceedings{zheng2023robustness,
  title={Robustness of Trajectory Prediction Models Under Map-Based Attacks},
  author={Zheng, Zhihao and Ying, Xiaowen and Yao, Zhen and Chuah, Mooi Choo},
  booktitle={Proceedings of the IEEE/CVF Winter Conference on Applications of Computer Vision},
  pages={4541--4550},
  year={2023}
}

@inproceedings{DBLP:conf/iccv/PellegriniESG09,
  author       = {Stefano Pellegrini and
                  Andreas Ess and
                  Konrad Schindler and
                  Luc Van Gool},
  title        = {You'll never walk alone: Modeling social behavior for multi-target
                  tracking},
  booktitle    = {{IEEE} 12th International Conference on Computer Vision, {ICCV} 2009,
                  Kyoto, Japan, September 27 - October 4, 2009},
  pages        = {261--268},
  year         = {2009},
  url          = {https://doi.org/10.1109/ICCV.2009.5459260},
  doi          = {10.1109/ICCV.2009.5459260},
  bibsource    = {dblp computer science bibliography, https://dblp.org}
}

@inproceedings{lerner2007crowds,
  title={Crowds by example},
  author={Lerner, Alon and Chrysanthou, Yiorgos and Lischinski, Dani},
  booktitle={Computer graphics forum},
  volume={26},
  pages={655--664},
  year={2007},
  organization={Wiley Online Library}
}

@article{madry2017towards,
  title={Towards deep learning models resistant to adversarial attacks},
  author={Madry, Aleksander and Makelov, Aleksandar and Schmidt, Ludwig and Tsipras, Dimitris and Vladu, Adrian},
  journal={arXiv preprint arXiv:1706.06083},
  year={2017}
}

@inproceedings{phan2020covernet,
  title={Covernet: Multimodal behavior prediction using trajectory sets},
  author={Phan-Minh, Tung and Grigore, Elena Corina and Boulton, Freddy A and Beijbom, Oscar and Wolff, Eric M},
  booktitle={Proceedings of the IEEE/CVF Conference on Computer Vision and Pattern Recognition},
  pages={14074--14083},
  year={2020}
}

@inproceedings{alahi2014socially,
  title={Socially-aware large-scale crowd forecasting},
  author={Alahi, Alexandre and Ramanathan, Vignesh and Fei-Fei, Li},
  booktitle={Proceedings of the IEEE Conference on Computer Vision and Pattern Recognition},
  pages={2203--2210},
  year={2014}
}

@inproceedings{robicquet2016learning,
  title={Learning social etiquette: Human trajectory understanding in crowded scenes},
  author={Robicquet, Alexandre and Sadeghian, Amir and Alahi, Alexandre and Savarese, Silvio},
  booktitle={Computer Vision--ECCV 2016: 14th European Conference, Amsterdam, The Netherlands, October 11-14, 2016, Proceedings, Part VIII 14},
  pages={549--565},
  year={2016},
  organization={Springer}
}

@inproceedings{cohen2019certified,
  title={Certified adversarial robustness via randomized smoothing},
  author={Cohen, Jeremy and Rosenfeld, Elan and Kolter, Zico},
  booktitle={international conference on machine learning},
  pages={1310--1320},
  year={2019},
  organization={PMLR}
}

@inproceedings{athalye2018obfuscated,
  title={Obfuscated gradients give a false sense of security: Circumventing defenses to adversarial examples},
  author={Athalye, Anish and Carlini, Nicholas and Wagner, David},
  booktitle={International Conference on Machine Learning},
  pages={274--283},
  year={2018},
  organization={PMLR}
}

@article{li2019certified,
  title={Certified adversarial robustness with additive noise},
  author={Li, Bai and Chen, Changyou and Wang, Wenlin and Carin, Lawrence},
  journal={Advances in neural information processing systems},
  volume={32},
  year={2019}
}

@inproceedings{lecuyer2019certified,
  title={Certified robustness to adversarial examples with differential privacy},
  author={Lecuyer, Mathias and Atlidakis, Vaggelis and Geambasu, Roxana and Hsu, Daniel and Jana, Suman},
  booktitle={2019 IEEE Symposium on Security and Privacy (SP)},
  pages={656--672},
  year={2019},
  organization={IEEE}
}

@article{velmurugan2010computational,
  title={Computational complexity between K-means and K-medoids clustering algorithms for normal and uniform distributions of data points},
  author={Velmurugan, T and Santhanam, T},
  journal={Journal of computer science},
  volume={6},
  number={3},
  pages={363},
  year={2010}
}

@InProceedings{bahari2025certified,
    author    = {Bahari, Mohammadhossein and Saadatnejad, Saeed and Askari Farsangi, Amirhossein and Moosavi-Dezfooli, Seyed-Mohsen and Alahi, Alexandre},
    title     = {Certified Human Trajectory Prediction},
    booktitle = {Proceedings of the IEEE/CVF Conference on Computer Vision and Pattern Recognition (CVPR)},
    year      = {2025},
}

@inproceedings{DBLP:conf/iclr/0003LCL25,
  author       = {Jianhua Sun and
                  Yuxuan Li and
                  Liang Chai and
                  Cewu Lu},
  title        = {Interactive Adjustment for Human Trajectory Prediction with Individual
                  Feedback},
  booktitle    = {The Thirteenth International Conference on Learning Representations,
                  {ICLR} 2025, Singapore, April 24-28, 2025},
  publisher    = {OpenReview.net},
  year         = {2025},
  url          = {https://openreview.net/forum?id=DCpukR83sw},
  bibsource    = {dblp computer science bibliography, https://dblp.org}
}

@inproceedings{DBLP:conf/iclr/HuC25,
  author       = {Bo Hu and
                  Tat{-}Jen Cham},
  title        = {TSC-Net: Prediction of Pedestrian Trajectories by Trajectory-Scene-Cell
                  Classification},
  booktitle    = {The Thirteenth International Conference on Learning Representations,
                  {ICLR} 2025, Singapore, April 24-28, 2025},
  publisher    = {OpenReview.net},
  year         = {2025},
  url          = {https://openreview.net/forum?id=Xmh5gdMfRJ},
  bibsource    = {dblp computer science bibliography, https://dblp.org}
}

@inproceedings{DBLP:conf/cvpr/JeongLPLY25,
  author       = {Jaewoo Jeong and
                  Seohee Lee and
                  Daehee Park and
                  Giwon Lee and
                  Kuk{-}Jin Yoon},
  title        = {Multi-modal Knowledge Distillation-based Human Trajectory Forecasting},
  booktitle    = {{IEEE/CVF} Conference on Computer Vision and Pattern Recognition,
                  {CVPR} 2025, Nashville, TN, USA, June 11-15, 2025},
  pages        = {24222--24233},
  publisher    = {Computer Vision Foundation / {IEEE}},
  year         = {2025},
  url          = {https://openaccess.thecvf.com/content/CVPR2025/html/Jeong\_Multi-modal\_Knowledge\_Distillation-based\_Human\_Trajectory\_Forecasting\_CVPR\_2025\_paper.html},
  bibsource    = {dblp computer science bibliography, https://dblp.org}
}

@inproceedings{DBLP:conf/cvpr/FuY0LL25,
  author       = {Yuxiang Fu and
                  Qi Yan and
                  Lele Wang and
                  Ke Li and
                  Renjie Liao},
  title        = {MoFlow: One-Step Flow Matching for Human Trajectory Forecasting via
                  Implicit Maximum Likelihood Estimation based Distillation},
  booktitle    = {{IEEE/CVF} Conference on Computer Vision and Pattern Recognition,
                  {CVPR} 2025, Nashville, TN, USA, June 11-15, 2025},
  pages        = {17282--17293},
  publisher    = {Computer Vision Foundation / {IEEE}},
  year         = {2025},
  url          = {https://openaccess.thecvf.com/content/CVPR2025/html/Fu\_MoFlow\_One-Step\_Flow\_Matching\_for\_Human\_Trajectory\_Forecasting\_via\_Implicit\_CVPR\_2025\_paper.html},
  bibsource    = {dblp computer science bibliography, https://dblp.org}
}

@inproceedings{DBLP:conf/cvpr/ChenZHFSWW25,
  author       = {Kai Chen and
                  Xiaodong Zhao and
                  Yujie Huang and
                  Guoyu Fang and
                  Xiao Song and
                  Ruiping Wang and
                  Ziyuan Wang},
  title        = {SocialMOIF: Multi-Order Intention Fusion for Pedestrian Trajectory
                  Prediction},
  booktitle    = {{IEEE/CVF} Conference on Computer Vision and Pattern Recognition,
                  {CVPR} 2025, Nashville, TN, USA, June 11-15, 2025},
  pages        = {22465--22475},
  publisher    = {Computer Vision Foundation / {IEEE}},
  year         = {2025},
  url          = {https://openaccess.thecvf.com/content/CVPR2025/html/Chen\_SocialMOIF\_Multi-Order\_Intention\_Fusion\_for\_Pedestrian\_Trajectory\_Prediction\_CVPR\_2025\_paper.html},
  bibsource    = {dblp computer science bibliography, https://dblp.org}
}

@inproceedings{DBLP:conf/cvpr/YuHWLWZ25,
  author       = {Yi Yu and
                  Weizhen Han and
                  Libing Wu and
                  Bingyi Liu and
                  Enshu Wang and
                  Zhuangzhuang Zhang},
  title        = {Enduring, Efficient and Robust Trajectory Prediction Attack in Autonomous
                  Driving via Optimization-Driven Multi-Frame Perturbation Framework},
  booktitle    = {{IEEE/CVF} Conference on Computer Vision and Pattern Recognition,
                  {CVPR} 2025, Nashville, TN, USA, June 11-15, 2025},
  pages        = {17229--17238},
  publisher    = {Computer Vision Foundation / {IEEE}},
  year         = {2025},
  url          = {https://openaccess.thecvf.com/content/CVPR2025/html/Yu\_Enduring\_Efficient\_and\_Robust\_Trajectory\_Prediction\_Attack\_in\_Autonomous\_Driving\_CVPR\_2025\_paper.html},
  bibsource    = {dblp computer science bibliography, https://dblp.org}
}

@article{jin2025reconcile,
  title={Reconcile Certified Robustness and Accuracy for DNN-based Smoothed Majority Vote Classifier},
  author={Jin, Gaojie and Yi, Xinping and Huang, Xiaowei},
  journal={arXiv preprint arXiv:2509.25979},
  year={2025}
}

@article{jin2025s,
  title={{S\textsuperscript{2}O}: Enhancing Adversarial Training with Second-Order Statistics of Weights},
  author={Jin, Gaojie and Yi, Xinping and Huang, Wei and Schewe, Sven and Huang, Xiaowei},
  journal={IEEE Transactions on Pattern Analysis and Machine Intelligence},
  year={2025},
  publisher={IEEE}
}

@article{wang2025clucert,
  title={CluCERT: Certifying LLM Robustness via Clustering-Guided Denoising Smoothing},
  author={Wang, Zixia and Jin, Gaojie and Hu, Jia and Mu, Ronghui},
  journal={arXiv preprint arXiv:2512.08967},
  year={2025}
}

\end{document}